\documentclass[10pt,twocolumn,letterpaper]{article}

\usepackage{cvpr}
\usepackage{times}
\usepackage{epsfig}
\usepackage{graphicx}
\usepackage{amsmath}
\usepackage{amssymb}

\usepackage{algorithmic}
\usepackage{cite}
\usepackage{array}
\usepackage{color}        
\usepackage{mdwmath}
\usepackage{mdwtab}
\usepackage[plain,noend]{algorithm2e}
\usepackage{tabularx, booktabs}
\usepackage{bigstrut}
\usepackage{bm}
\usepackage{epsfig}
\usepackage{subfigure}
\usepackage{wrapfig}
\usepackage{microtype}
\usepackage{multirow}
\usepackage{enumitem}
\usepackage{eso-pic}
\usepackage{xspace}
\def\secvspace{\vspace{-2mm}}
\def\eqnvspace{\vspace{-2mm}}
\def\figvspace{\vspace{-2mm}}

\newcommand{\Paragraph}[1]{\vspace{0mm} \noindent \textbf{#1} \hspace{0mm}}

\newcommand{\xiyin}[1]{\textcolor{black}{#1}}

\usepackage{enumitem}
\setitemize[0]{leftmargin=10pt}

\newenvironment{tight_itemize}{
\begin{itemize}[leftmargin=10pt]
  \setlength{\topsep}{0pt}
  \setlength{\itemsep}{2pt}
  \setlength{\parskip}{0pt}
  \setlength{\parsep}{0pt}
}{\end{itemize}}

\usepackage[pagebackref=true,breaklinks=true,letterpaper=true,colorlinks,bookmarks=false]{hyperref}

\cvprfinalcopy 

\def\cvprPaperID{3198}

\ifcvprfinal\fi
\begin{document}

\title{Feature Transfer Learning for Face Recognition with Under-Represented Data}

\author{Xi Yin$^{\dagger}$\thanks{Main part of the work is done when Xi was an intern at NEC Laboratories America.}, Xiang Yu$^{\ddag}$, Kihyuk Sohn$^{\ddag}$, Xiaoming Liu$^{\dagger}$ and Manmohan Chandraker$^{\S\ddag}$\\
$^{\dagger}$Michigan State University \\
$^{\ddag}$ NEC Laboratories America \\
$^{\S}$University of California, San Diego \\
{\tt\small \{yinxi1,liuxm\}@cse.msu.edu, \{xiangyu,ksohn,manu\}@nec-labs.com}}

\maketitle

\begin{abstract}
\secvspace
Despite the large volume of face recognition datasets, there is a significant portion of subjects, of which the samples are insufficient and thus under-represented. Ignoring such significant portion results in insufficient training data. Training with under-represented data leads to biased classifiers in conventionally-trained deep networks. In this paper, we propose a center-based feature transfer framework to augment the feature space of under-represented subjects from the regular subjects that have sufficiently diverse samples. A Gaussian prior of the variance is assumed across all subjects and the variance from regular ones are transferred to the under-represented ones. This encourages the under-represented distribution to be closer to the regular distribution. Further, an alternating training regimen is proposed to simultaneously achieve less biased classifiers and a more discriminative feature representation. We conduct ablative study to mimic the under-represented datasets by varying the portion of under-represented classes on the MS-Celeb-1M dataset. Advantageous results on LFW, IJB-A and MS-Celeb-1M demonstrate the effectiveness of our feature transfer and training strategy, compared to both general baselines and state-of-the-art methods. Moreover, our feature transfer successfully presents smooth visual interpolation, which conducts disentanglement to preserve identity of a class while augmenting its feature space with non-identity variations such as pose and lighting.
\end{abstract}

\vspace{-2mm}
\section{Introduction}
Face recognition is one of the ongoing success stories in the deep learning era, yielding very high accuracy on several benchmarks ~\cite{lfwdatabase,ijba,guo2016msceleb}. However, it remains undetermined how deep learning classifiers for fine-grained recognition are trained to maximally exploit real-world data. While it is known that recognition engines are data-hungry and keep improving with more volume, mechanisms to derive benefits from the vast diverse data are relatively unexplored. In particular, as discussed by~\cite{vanhorn2017}, there is a non-negligible part of data that is under-represented (\textit{UR}), where only a few samples are available for each class. 

\begin{figure}[!!t]
\begin{center}
\begin{tabular}{@{}c@{\hspace{2mm}}c@{}}
\includegraphics[width=0.23\textwidth]{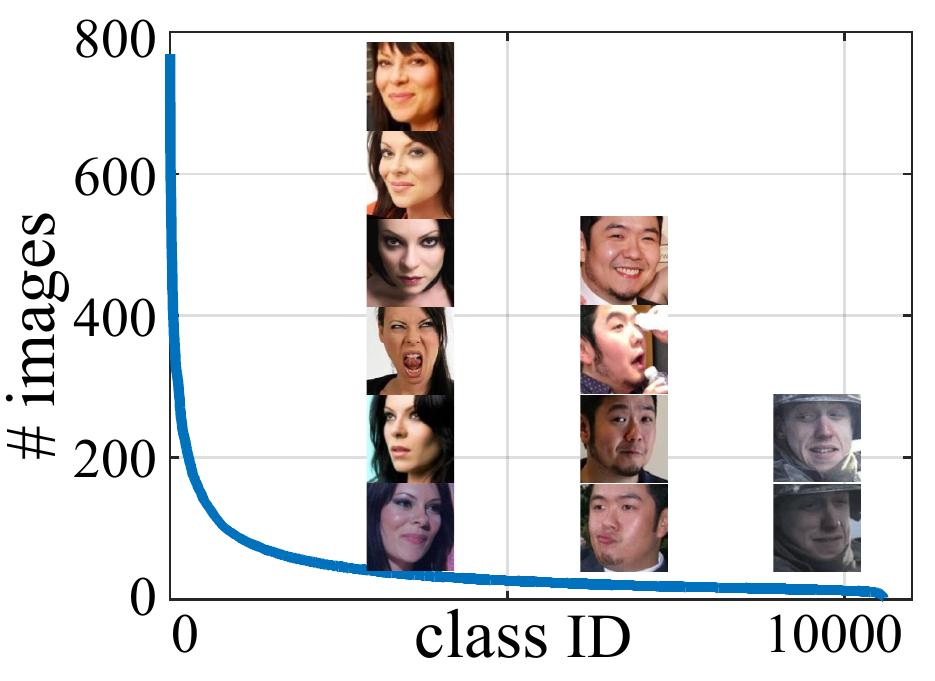} &
\includegraphics[width=0.23\textwidth]{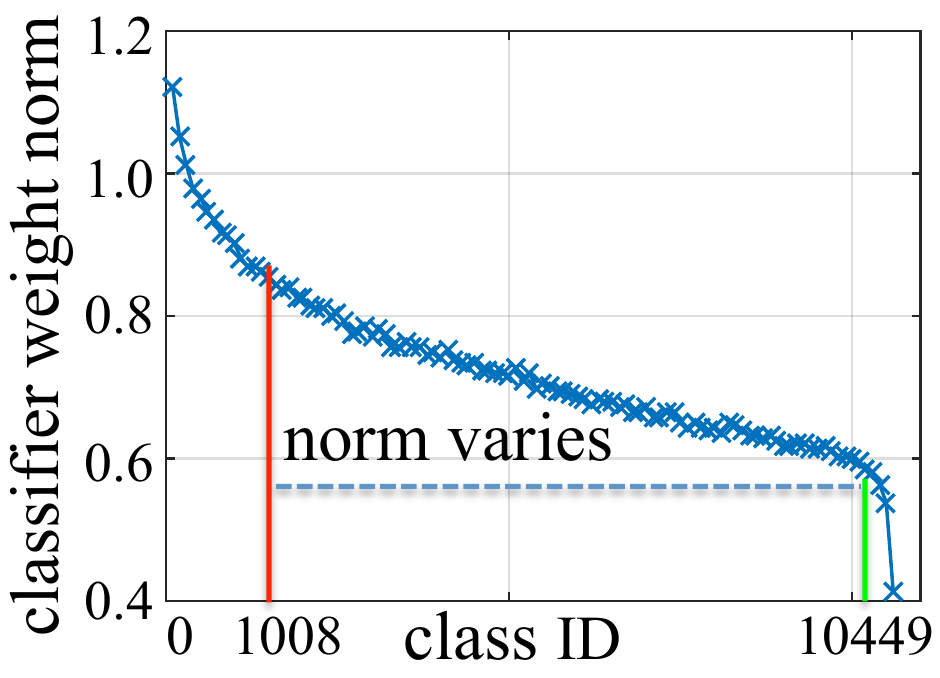} \\ [-1mm]
\small (a) & (b) \\ [-1mm]
\includegraphics[trim=0 5 430 5, clip,width=0.21\textwidth]{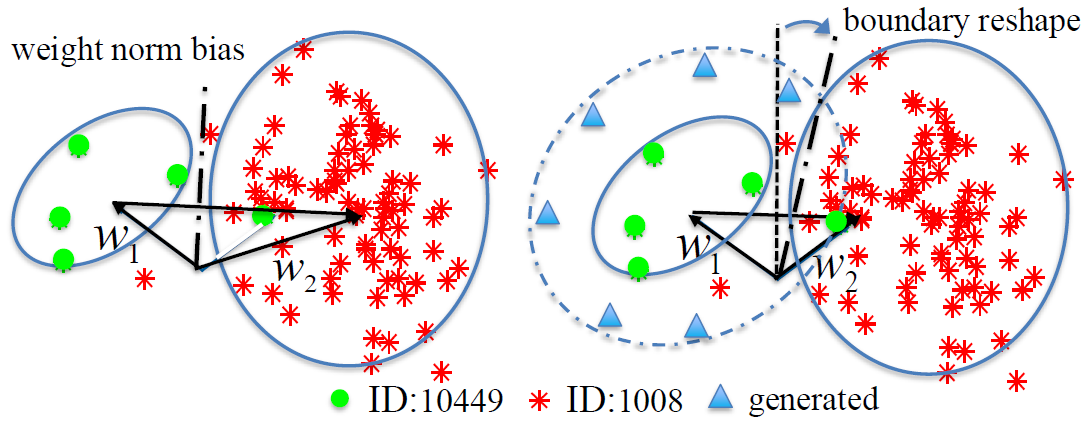} &
\includegraphics[trim=390 5 5 5, clip,width=0.23\textwidth]{fig/casia_exp_cd.PNG} \\ [-1mm]
\small (c) & (d) \\ [-1mm]
\end{tabular}
\end{center}
\figvspace
\caption{Illustration of the {\textit{UR}} data problem and our proposed solution. (a) The data distribution of CASIA-WebFace dataset~\cite{yi2014learning}. (b) Classifier weight norm varies across classes in proportion to their volume. (c) Weight norm for regular class $1008$ is larger than \textit{UR} class $10449$, causing a bias in the decision boundary (dashed line) towards ID $10449$. (d) Data re-sampling solves the classifier bias to some extent. However, the variance of ID 1008 is much larger than ID $10449$. We augment the feature space of ID $1008$ (dashed ellipsoid) and propose improved training strategies, which corrects the classifier bias and learns a better feature representation.}
\label{fig:casia_exp}
\vspace{-3mm}
\end{figure}

It is evident that classifiers that ignore this \textit{UR} data likely imbibe hidden biases. Consider CASIA-Webface~\cite{yi2014learning} dataset as an example (Figure~\ref{fig:casia_exp} (a)). About $39\%$ of the $10$K subjects have less than $20$ images. A simple solution is to discard the \textit{UR} classes, which results in insufficient training data. 
Besides reduction in the volume of data, the inherently uneven sampling leads to bias in the weight norm distribution across regular and \textit{UR} classes (Figure~\ref{fig:casia_exp} (b,c)). Sampling \textit{UR} classes at a higher frequency alleviates the problem, but still leads to biased decision boundaries due to insufficient intra-class variance in \textit{UR} classes (Figure~\ref{fig:casia_exp} (d)).

In this paper, we propose \xiyin{Feature Transfer Learning (FTL)} to train less biased face recognition classifiers by adapting the feature distribution of \textit{UR} classes to mimic that of regular classes. Our FTL handles such \textit{UR} classes during training by augmenting their feature space using a center-based transfer. 
In particular, assuming a Gaussian prior on features with class-specific mean and the shared variance across regular and \textit{UR} classes, we generate new samples of \textit{UR} classes at feature space, by transferring the linear combination of the principal components of variance that are estimated from regular classes to the \textit{UR} classes. 

Our feature transfer addresses the issue of imbalanced training data. However, using the transferred data directly for training is sub-optimal as the transfer might skew the class distributions. Thus, we propose a training regimen that alternates between carefully designed choices to solve for feature transfer (with the goal of obtaining a less biased decision boundary) and feature learning (with the goal of learning a more discriminative representation) simultaneously. 
Besides, we propose a novel and effective metric regularization which contributes to the general deep training in an orthogonal way.

To study the empirical properties of our method, we construct \textit{UR} datasets by limiting the number of samples for various proportions of classes in MS-Celeb-1M~\cite{guo2016msceleb}, and evaluate on LFW~\cite{lfwdatabase}, IJB-A~\cite{ijba} and the hold-out test set from MS-Celeb-1M. We observe that our FTL consistently improves upon baseline method that does not specifically handle \textit{UR} classes. 
Advantageous results over state-of-the-art methods on LFW and IJB-A further confirm the effectiveness of the feature transfer module. Moreover, our FTL can be applied to low-shot or one-shot scenarios, where a few samples are available for some classes. Competitive record on MS-celeb-1M one-shot challenge~\cite{guo2017one} evidences the advantage. Finally, we visualize our feature transfer module through smooth feature interpolation. It shows that for our feature representation, identity is preserved while non-identity aspects are successfully disentangled and transferred to the target subject.

\begin{figure*}[t]
\centering
\includegraphics[trim=15 230 25 185, clip, width=0.99\textwidth]{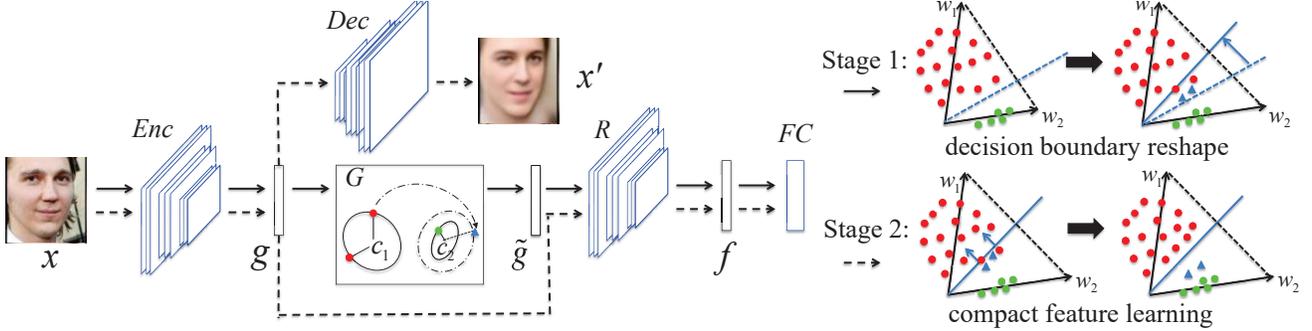}
\vspace{-2mm}
\caption{Overview of our proposed FTL framework. It consists of a feature extractor $Enc$, a decoder $Dec$, a feature filter $R$, a fully connected layer as classifier $FC$, and a feature transfer module $G$. The network is trained with an alternating bi-stage strategy. At stage $1$ (solid arrows), we fix $Enc$ and apply feature transfer $G$ to generate new feature samples (blue triangles) that are more diverse to reshape the decision boundary. In stage $2$ (dashed arrows), we fix the rectified classifier $FC$, and update all the other models. As a result, the samples that are originally on or across the boundary are pushed towards their center (blue arrows in bottom right). Best viewed in color.}
\label{fig:overview}
\vspace{-3mm}
\end{figure*}

We summarize our contributions as the following items.
\vspace{-2mm}
\begin{tight_itemize}
\item{A center-based feature transfer algorithm to enrich the distribution of \textit{UR} classes, leading to diversity without sacrificing volume. It also leads to an effective disentanglement of identity and non-identity representations.}
\item{A two-stage alternative training scheme to achieve a less biased classifier and retain discriminative power of the feature representation.}
\item{A simple but effective metric regularization to enhance performance for both our method and baselines, which is also applicable to other recognition tasks.}
\item{Extensive ablation experiments demonstrate the effectiveness of our FTL framework. Combining with the proposed m-L2 regularization and other orthogonal metric learning methods, we achieve top performance on LFW and IJB-A.}
\end{tight_itemize}

\section{Related Work}
\noindent\textbf{Imbalanced data classification}
Classic works study data re-sampling methods~\cite{smote2002,he2009}, which learn unbiased classifiers by changing the sampling frequency.
By applying deep neural networks~\cite{alex2012,resnet}, the frontier of face recognition research has been significantly advanced \cite{schroff2015facenet,wen2016discriminative,liu2017sphereface}. 
However, there are only few works that discuss about learning from \textit{UR} data.
Huang \etal~\cite{chenhuang2016} propose quintuplet sampling based hinge loss to maintain both inter-cluster and inter-class margins. 
Zhang \etal~\cite{zhang2017range} propose the range loss that simultaneously reduces intra-class variance and enlarges the inter-class variance. 
However, {\textit{UR}} classes are treated in the same way as regular classes in the above methods. 
Guo and Zhang~\cite{guo2017one} propose \textit{UR} class promotion loss that regularizes the norm of weight vectors of \textit{UR} classes, which can solve the unbalance issue to some extent. 
Other than designing data sampling rules or regularization on \textit{UR} classes, we augment \textit{UR} classes by generating feature-level samples through transfer of intra-class variance from regular classes, \xiyin{which solves the fundamental problem of {\textit{UR}} data.}

\vspace{-1mm}
\noindent\textbf{One-shot and low-shot learning}
Low-shot learning aims at recognizing an image for a specific class with very few or even one image available at training. Some efforts are made by enforcing strong regularization \cite{hariharan2017} or utilizing non-parametric classification methods based on distance metric learning \cite{vinyals2016,NIPS2016_6200}. Generative model based methods have also been studied in recent years. Dixit \etal~\cite{dixit2017} propose a data augmentation method using attribute-guided feature descriptor for generation. The method in~\cite{hariharan2017} proposes non-parametric generation of features by transferring within class pair-wise variation from regular classes in object classification task. Compared to their task on ImageNet~\cite{ILSVRC15} with $1$K classes, face recognition is a fine-grained classification problem that incorporates at least two orders of magnitude more classes with low inter-class variance.

\vspace{-1mm}
\noindent\textbf{Feature transfer learning}
Transfer learning applies information from a known domain to an unknown one~\cite{learning-person-specific-models-for-facial-expression-and-action-unit-recognition,transfer-learning-with-one-class-data}. We refer to~\cite{sinno2009} for further discussion. Attributes are used in~\cite{dixit2017} to synthesize feature-level data. In~\cite{ksohn_video_recognition_2017}, features are transferred from web images to video frames via a generative adversarial network (GAN)~\cite{goodfellow2014generative}. Our method shares the same flavor in terms of feature transfer concept. However, compared to~\cite{ksohn_video_recognition_2017}, no additional supervision is provided in our method as it may introduce new bias. We model the intra-class variance in a parametric way, assuming the regular classes and \textit{UR} classes share the same feature variance distribution. By transferring this shared variance, we transfer sample features from regular classes to \textit{UR} classes.

\secvspace
\section{The Proposed Approach}
In this section, we first introduce the problems caused by training with \textit{UR} classes for face recognition (Sec.~\ref{sec:3.1}). 
Then, we present the recognition backbone framework with our proposed metric regularization (Sec.~\ref{sec:3.2}), our proposed feature transfer framework (Sec.~\ref{sec:3.3}), and the alternating training scheme to solve these problems (Sec.~\ref{sec:3.4}). 

\begin{figure*}[t]
\centering
\includegraphics[width=0.98\textwidth]{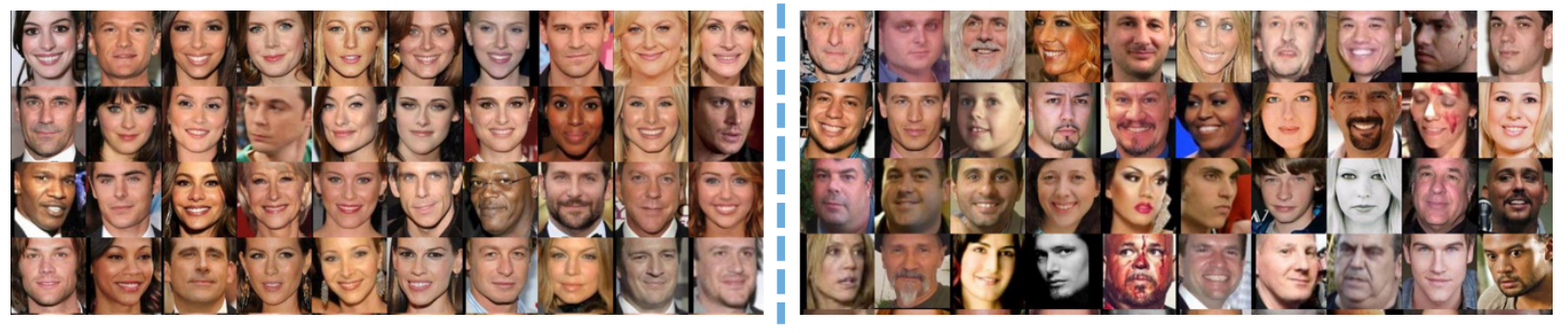}
\vspace{-2mm}
\caption{Visualization of samples closest to the feature center of classes with most number of images (left) and classes with least number of images (right). We find that near-frontal close-to-neutral faces are the nearest neighbors of the feature centers of regular classes. However, the nearest neighbors of the feature centers of \textit{UR} classes still contain pose and expression variations. Features are extracted by VGGFace model~\cite{ParkhiVGG} and samples are from CASIA-WebFace dataset.}
\label{fig:centerface}
\vspace{-4mm}
\end{figure*}

\subsection{Limitations of Training with \textit{UR} Classes}
\label{sec:3.1}
A recent work~\cite{zhang2017range} shows that directly learning face representation with \textit{UR} classes results in degraded performance. To demonstrate the problems of training with \textit{UR} classes, we train a network (CASIA-Net) on CASIA-Webface~\cite{yi2014learning}, of which the data distribution is shown in Figure~\ref{fig:casia_exp} (a).
We mainly observe two issues: (1) wildly variant classifier weight norms; and (2) imbalanced intra-class variances between regular and \textit{UR} classes. 

\Paragraph{Imbalance on classifier weight norm}
As shown in Figure~\ref{fig:casia_exp} (b), the norms of the classifier weights (\ie, the weights in the last fully connected layer) of regular classes are much larger than those of \textit{UR} classes, which causes the decision boundary biases towards the \textit{UR} classes~\cite{guo2017one}. This is because the much larger volume of regular classes lead to more frequent weight updates than those of \textit{UR} classes. To alleviate this problem, there are typical solutions such as data re-sampling or weight normalization~\cite{guo2017one}. However, such strategies can not solve the fundamental problem of lacking sufficient and diversified samples in \textit{UR} classes, which is demonstrated in the following.

\Paragraph{Imbalance on intra-class variance}
As an illustrative example, we randomly pick two classes, one regular class (ID=$1008$) and one \textit{UR} class (ID=$10449$). We visualize the features from two classes projected onto 2D space using t-SNE~\cite{tsne_maaten} in Figure~\ref{fig:casia_exp}(c). Further, the feature space after weight norm regularization is shown in Figure~\ref{fig:casia_exp}(d). Although the weight norms are regularized to be similar, the low intra-class variance of the \textit{UR} class still causes the decision boundary bias problem. 

Based on these observations, we posit that \emph{enlarging the intra-class variance for \textit{UR} classes is the key to alleviate these imbalance issues}. Therefore, we propose a feature transfer learning approach that generates extra samples for \textit{UR} classes to enlarge the intra-class variance. As illustrated in Figure~\ref{fig:casia_exp}(d), the feature distribution augmented by the virtual samples (blue triangles) helps to rectify the classifier decision boundary and learn a better representation.

\subsection{The Proposed Framework}
\label{sec:3.2}
Most recent success in deep face recognition works on novel losses or regularizations~\cite{sankaranarayanan2016triplet,liu2017sphereface,crosswhite2017template,schroff2015facenet,NIPS2016_6200}, which aim at improving model generalization. In contrast, our method focuses on enlarging intra-class variance of \textit{UR} classes by transferring knowledge from regular classes.
At first glance, our goal of diversifying features seems to contradict with the general premise of face recognition frameworks, \ie, pursuing compact features. 
In fact, we enlarge the intra-class variance of \textit{UR} classes at a lower level feature space, which we term as rich-feature layer~\cite{rich_feature}. The subsequent filtering layers will learn a more discriminative representation.

As illustrated in Figure~\ref{fig:overview}, the proposed framework is composed of several modules including an encoder, decoder, feature transfer module followed by filtering module and a classifier layer. 
An encoder $Enc$ computes rich features ${\bf{g}}=Enc({\bf{x}}) \in \mathbb{R}^{320}$ from an input image ${\bf{x}} \in \mathbb{R}^{100\times 100}$ and reconstructs the input with a decoder $Dec$, \ie, ${\bf{x'}}=Dec({\bf{g}})=Dec(Enc({\bf{x}})) \in \mathbb{R}^{100\times 100}$. 
This pathway is trained with the following pixel-wise reconstruction loss:
\eqnvspace
\begin{equation}
\mathcal{L}_{recon} = \|{\bf{x'}} - {\bf{x}}\|_2^2.
\label{eq:recon}
\eqnvspace
\end{equation}

The reconstruction loss allows ${\bf{g}}$ to contain diverse non-identity variations such as pose, expression, and lighting. Therefore, we denote $\bf{g}$ as the rich feature space. 

A filtering network $R$ is applied to generate discriminative identity features ${\bf{f}}=R(\textbf{g}) \in \mathbb{R}^{320}$ that are fed to a linear classifier layer $FC$ with weight matrix ${\bf{W}}=[{\bf{w}_{j}}]_{j=1}^{N_c}\in\mathbb{R}^{N_{c}\times 320}$ where $N_c$ is the total number of classes. 
This pathway optimizes the softmax loss:
\eqnvspace
\begin{equation}
\mathcal{L}_{sfmx} =- \log \frac{\exp({\bf{w}}^T_{y_i} {\bf{f}})}{\sum_{j}^{N_c} \exp({\bf{w}}^T_j {\bf{f}})},
\label{eq:classifier}
\eqnvspace
\end{equation}
\xiyin{where $y_i$ is the ground-truth identity label of $\bf{x}$}.

Note that softmax loss is \emph{scale-dependent} where the loss can be made arbitrarily small by scaling the norm of the weights ${\bf{w}}_{j}$ or features ${\bf{f}}$. Typical solutions to prevent this problem are to either regularize the norm of weights\footnote{\url{http://ufldl.stanford.edu/wiki/index.php/Softmax_Regression\#Weight_Decay}} or features, or to normalize both of them~\cite{wang2017normface}. However, we argue that these methods are too stringent since they penalize norms of individual weights and features without considering their compatibility.
Instead, we propose to regularize the norm of the output of $FC$ as following:
\eqnvspace
\begin{equation}
\mathcal{L}_{reg} = \|{\bf{W}}^T{\bf{f}}\|_2^2.
\label{eq:ml2}
\eqnvspace
\end{equation}

We term the proposed regularization as metric $L_2$ or m-$L_2$ regularization.
As will shown in the experiment, joint regularization on weights and features works better than individual regularization.

Finally, we formulate the training loss in Eqn.~\eqref{eq:loss_sum}, with the following coefficients $\alpha_{sfmx}\,{=}\,\alpha_{recon}\,{=}\,1,\alpha_{reg}\,{=}\,0.25$ unless otherwise stated:
\eqnvspace
\begin{equation}
\mathcal{L} = \alpha_{sfmx} \mathcal{L}_{sfmx} + \alpha_{recon} \mathcal{L}_{recon} + \alpha_{reg} \mathcal{L}_{reg}.
\label{eq:loss_sum}
\eqnvspace
\end{equation}

\secvspace
\subsection{Feature Transfer for \textit{UR} Classes}
\label{sec:3.3}
Following the Joint Bayesian face model~\cite{chen2012bayesian}, we assume that the rich feature ${\bf{g}}_{ik}$ from class $i$ lies in a Gaussian distribution with a class mean ${\bf{c}}_{i}$ and a covariance matrix $\Sigma_{i}$.
The class mean or center is estimated as an arithmetic average over all features from the same class. As shown in the left of Figure~\ref{fig:centerface}, the center representation of regular classes is identity-specific while removing non-identity factors such as pose, expression and illumination. However, as in the right of Figure~\ref{fig:centerface}, due to the lack of samples, the center estimation of \textit{UR} classes is not accurate and often biased towards certain identity-irrelevant factors like pose, which we find to be dominant in practice.

To improve the quality of center estimation for \textit{UR} classes, we discard samples with extreme pose variation. Furthermore, we consider averaging features from both the original and horizontally flipped images. With $\bar{\bf{g}}_{ik}\in\mathbb{R}^{320}$ denoting the rich feature extracted from the flipped image, the feature center is estimated as follows:
\eqnvspace
\begin{equation}
{\bf{c}}_{i}{=}\frac{1}{2|\Omega_{i}|}\sum_{k \in \Omega_{i}}({\bf{g}}_{ik}{+}\bar{\bf{g}}_{ik}),\, \Omega_{i}{=}\{k \: | \: |p_{ik}| {+} |\bar{p}_{ik}| \,{\le}\, \tau\},
\label{eq:center}
\eqnvspace
\end{equation}
\xiyin{where $p_{ik}$ and $\bar{p}_{ik}$ are the estimated poses of the original and flipped images, respectively. 
By bounding the summation, we expect the yaw angle $p_{ik}$ to be an inlier. }

To transfer the intra-class variance from regular classes to \textit{UR} classes, we assume the covariance matrices are shared across all classes, i.e., ${\bf{\Sigma}}_{i}\,{=}\,\bf{\Sigma}$. In theory, one can draw feature samples of \textit{UR} classes by adding a noise vector $\bf{\epsilon}\,{\sim}\,\mathcal{N}({\bf{0}},\bf{\Sigma})$ to its center ${\bf{c}}_{i}$. However, the direction of the noise vector might be too random and does not reflect the true factors of variations found in the regular classes. Therefore, we transfer the intra-class variance evaluated from the samples of regular classes. 
\xiyin{First, we calculate the covariance matrix $\bf{V}$ via:
\eqnvspace
\begin{equation}
    {\bf{V}} = \sum_{i=1}^{N_c} \sum_{k=1}^{m_i} ({\bf{g}}_{ik} - {\bf{c}}_{i})^{T}({\bf{g}}_{ik} - {\bf{c}}_{i})
\eqnvspace
\end{equation}
where $m_i$ is the total number of samples for class $i$. We perform PCA to decompose $\bf{V}$ into major components and take the first $150$ Eigenvectors as ${\bf{Q}}\in\mathbb{R}^{320\times150}$, which preserves $95\%$ energy.}
Our center-based feature transfer is achieved via:
\eqnvspace
\begin{equation}
\tilde{\bf{g}}_{ik} = {\bf{c}}_{i} + \mathbf{Q}\mathbf{Q}^{T}({\bf{g}}_{jk} - {\bf{c}}_{j}),
\label{eq:linearTransfer}
\eqnvspace
\end{equation}
where ${\bf{g}}_{jk}$ and ${\bf{c}}_j$ are the feature-level sample and the center of a regular class $j$. ${\bf{c}}_i$ is the feature center of an \textit{UR} class $i$ and $\tilde{\bf{g}}_{ik}$ is the transferred features for class $i$. Here, $\tilde{\bf{g}}_{ik}$ preserves the same identity as ${\bf{c}}_i$, with similar intra-class variance as ${\bf{g}}_{jk}$. By sufficiently sampling ${\bf{g}}_{jk}$ across different regular classes, we expect to obtain an enriched distribution of the \textit{UR} class $i$, which consists of both the original features ${\bf{g}}_{ik}$ and the transferred features $\tilde{\bf{g}}_{ik}$.

\subsection{Alternating Training Strategy}
\label{sec:3.4}
Given a training set of both regular and \textit{UR} classes $\mathbb{D} = \{ \mathbb{D}_{reg}, \mathbb{D}_{UR}\}$, we first pre-train all modules $\mathbb{M} = \{Enc, Dec, R, FC\}$ using Eqn.~\ref{eq:loss_sum} without feature transfer. 
Then, we alternate between the training of the classifier with our proposed feature transfer method for decision boundary reshape and learning a more discriminative feature representation with boundary-corrected classifier. The overview of our two-stage alternating training process is illustrated in Algorithm~\ref{alg:alternative}, which we describe in more details below.
\begin{algorithm}[t]
\caption{Two-stage alternating training strategy.} 
\label{alg:alternative}
\DontPrintSemicolon{
\SetAlCapFnt{\small}
\hrulefill \\
{\bf Stage $\bf{1}$: Decision boundary reshape.} \\
\hskip 0.5em Fixed models: $Enc$ and $Dec$. \\
\hskip 0.5em Training models: $R$ and $FC$, using Eqn.~\ref{eq:classifier} and ~\ref{eq:ml2}. \\ 
\hskip 0.5em Init $[{\bf{C}}, {\bf{Q}}, {\bf{h}}]$ = \textbf{\textit{{\textcolor{red}{UpdateStats}}}}(), $N_{iter}=\#$ iterations. \\
\hskip 0.5em \For{$i=1, \dots, N_{iter}$}
{
Train $1$st batch sampled from ${\bf{h}}$ in $\mathbb{D}_{reg}$: $\{{\bf{x}}^r, {\bf{y}}^r\}$. \\
Train $2$nd batch sampled from $\mathbb{D}_{UR}$: $\{{\bf{x}}^u, {\bf{y}}^u\}$. \\ 
Feature transfer: $\tilde{\bf{g}}^u$ = \textbf{\textit{{\textcolor{red}{Transfer}}}}(${\bf{x}}^r$, ${\bf{y}}^r$, ${\bf{y}}^u$). \\
Train $3$rd batch: $\{\tilde{\bf{g}}^u, {\bf{y}}^u\}$.
}
{\bf Stage $\bf{2}$: Compact feature learning.} \\
\hskip 0.5em Fixed models: $FC$. \\
\hskip 0.5em Training models: $Enc$, $Dec$, and $R$, using Eqn.~\ref{eq:loss_sum}. \\ 
\hskip 0.5em \For{$i=1, \dots, N_{iter}$}
{
train batch sampled from $\mathbb{D}$: $\{{\bf{x}}, {\bf{y}}\}$. 
}
Alternate stage $1$ and $2$ every $N_{iter}$ until convergence. \\  \hrulefill

{\bf{Function}} $[{\bf{C}}, {\bf{Q}}, {\bf{h}}]$ = \textbf{\textit{{\textcolor{red}{UpdateStats}}}}() \\
\renewcommand{\baselinestretch}{1.0}
\SetAlCapFnt{\small}
Init ${\bf{C}}=[]$, $\bf{V}=[]$, $\bf{h}=[]$, $m_{i}=\#$samples in class $i$,\\
$N_{c}=\#$ classes, $N_{s}=\#$ samples in each batch. \\
\For{$i=1, \dots, N_c$}
{
\For{$j=1, \dots, m_i$}{
${\bf{g}}_{ij}$ = $Enc({\bf{x}}_{ij})$, $\bar{\bf{g}}_{ij}$ = $Enc(\bar{\bf{x}}_{ij})$ \\
}
${\bf{c}}_{i}=\frac{1}{2|\Omega_{i}|}\sum_{k \in \Omega_{i}}({\bf{g}}_{ik}+\bar{\bf{g}}_{ik})$ \\
${\bf{C}}$.append(${\bf{c}}_i$) \\
\If{$i$ in $\mathbb{D}_{reg}$}
{
$d_i = \frac{1}{m_i}\sum_k{||{\bf{g}}_{ik} - {\bf{c}}_i ||_2}$ \\
\For{$j=1, \dots, m_i$}
{
$\bf{V}$ += $({\bf{g}}_{ij} - {\bf{c}}_i)^T({\bf{g}}_{ij} - {\bf{c}}_i)$ \\
\If{$||{\bf{g}}_{ij} - {\bf{c}}_i ||_2 > d_i$}
{$\bf{h}$.append([i,j])}
}
}
}
${\bf{Q}} = $ PCA($\bf{V}$) \\
{\bf{Function}} $\tilde{\bf{g}}^u$ = \textbf{\textit{{\textcolor{red}{Transfer}}}}(${\bf{x}}^r$, ${\bf{y}}^r$, ${\bf{y}}^u$) \\
${\bf{g}}^{r}$ = $Enc({\bf{x}}^{r})$ \\
\For{$k=1, \dots, N_s$}
{
${\bf{c}}_j = {\bf{C}}({\bf{y}}^r_k, :)$, ${\bf{c}}_i = {\bf{C}}({\bf{y}}^u_k, :)$ \\
$\tilde{\bf{g}}^u_k = {\bf{c}}_{i} + \mathbf{Q}\mathbf{Q}^{T}({\bf{g}}^r_{k} - {\bf{c}}_{j})$ \\ \hrulefill
}
}
\end{algorithm}

\Paragraph{Stage $1$: Decision boundary reshape.} 
In this stage, we train $R$ and $FC$ while fixing other modules (the rich feature space is fixed for stable feature transfer).
The goal is to reshape the decision boundary by transferring features from regular classes to \textit{UR} classes.
We first update the statistics for each regular class including the feature centers $\bf{C}$, PCA basis $\bf{Q}$ and an index list $\bf{h}$ of hard samples whose distances to the feature centers exceeding the average distance.
The PCA basis $\bf{Q}$ is achieved by decomposing the covariance matrix $\bf{V}$ computed with the samples from all regular classes $\mathbb{D}_{reg}$.
Three batches are applied for training in each iteration: (1) a regular batch sampled from hard index list $\bf{h}$: $\{{\bf{g}}^r, {\bf{y}}^r\}$, to guarantee no degradation in the performance; (2) a \textit{UR} batch sampled from \textit{UR} classes $\{{\bf{g}}^u, {\bf{y}}^u\}$, to conduct the updating similar to class-balanced sampling; (3) a transferred batch $\{\tilde{\bf{g}}^u,{\bf{y}}^u\}$ by transferring the variances from regular batch to \textit{UR} batch, to reshape the decision boundary. 

\Paragraph{Stage $2$: Compact feature learning.} 
In this stage, we train $Enc$, $Dec$ and $R$ using normal batches $\{{\bf{x}}, {\bf{y}}\}$ from both regular and \textit{UR} classes without feature transfer. We keep $FC$ fixed since it is already updated from the previous stage with decision boundary correction. The gradient directly back-propagates to $R$ and $Enc$ to learn a more compact representation that reduces the violation of crossing rectified classifier boundaries. 
We perform online alternation between stage $1$ and $2$ for every $N_{iter}$ iterations until convergence.

\section{Experiments}
We use MS-Celeb-1M as our training set.
Due to label noise, we adopt a cleaned version from~\cite{wu2015light} and remove the classes overlapped with LFW and IJB-A, which results in $4.8$M images of $76.5$K classes.
A class with no more than $20$ images is considered as a \textit{UR} class, following~\cite{zhang2017range}. A facial key point localization method~\cite{xiang_ddn_2016} is applied as the face alignment and cropping.

We apply an encoder-decoder structure for model $Enc$ and $Dec$. 
Model $R$ consists of a linear layer, two de-convolution layers, two convolution layers and another linear layer to obtain ${\bf{f}}\in\mathbb{R}^{320}$. Detail of the network structure is referred to the supplementary material. Adam solver with a learning rate of $2e^{-4}$ is used in model pre-training. A learning rate of $1e^{-5}$ is used in stage $1$ and $2$, which alternate for every $5$K iterations until convergence. The hyper-parameter setting is determined by an off-line parameter search based on a hold-out validation set.

\begin{figure}[!!t]
\begin{center}
\begin{tabular}{@{}c@{\hskip 1mm}c@{}}
\includegraphics[trim=25 0 40 10, clip, width=0.25\textwidth]{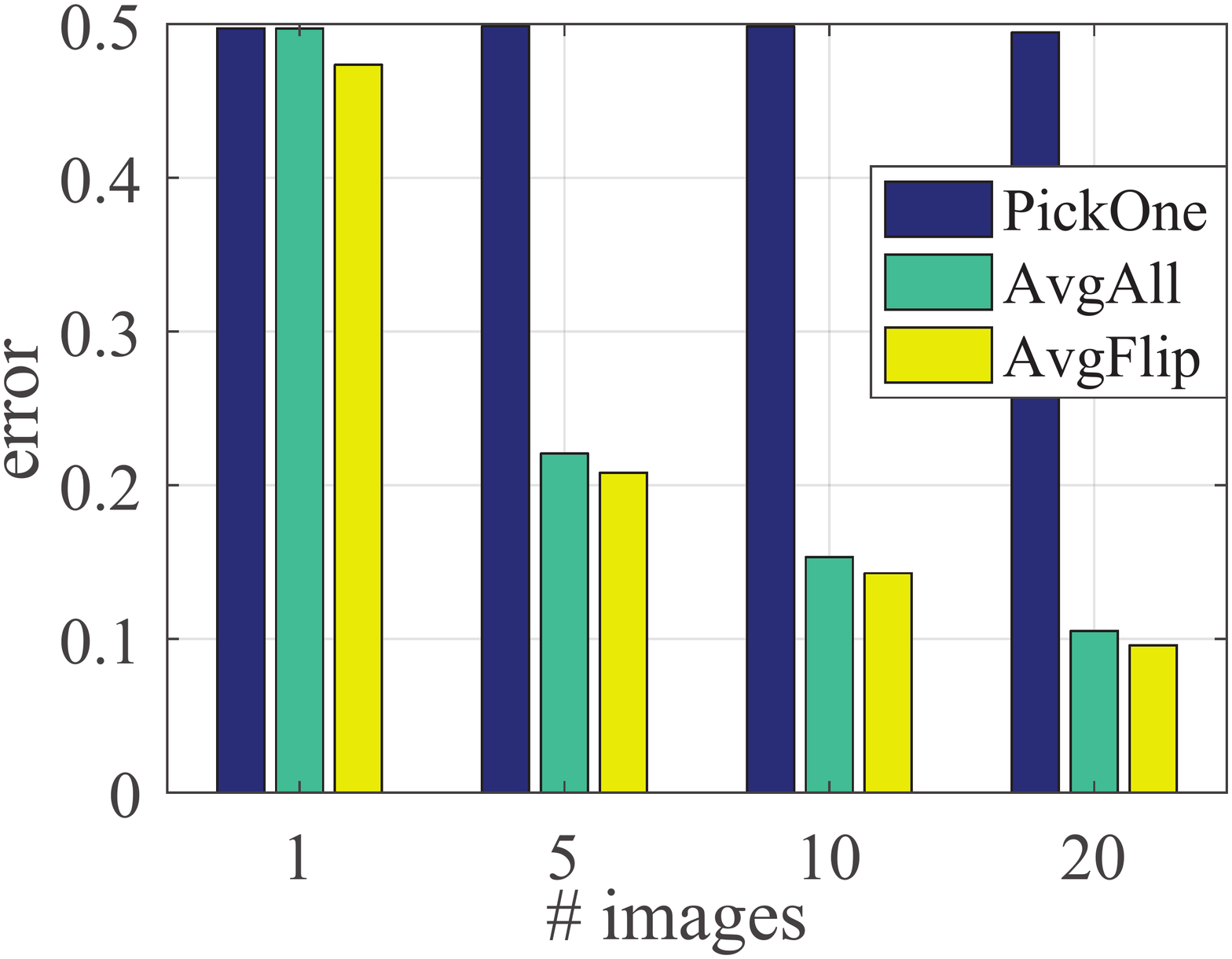} &
\includegraphics[trim=50 30 40 0, clip, width=0.24\textwidth]{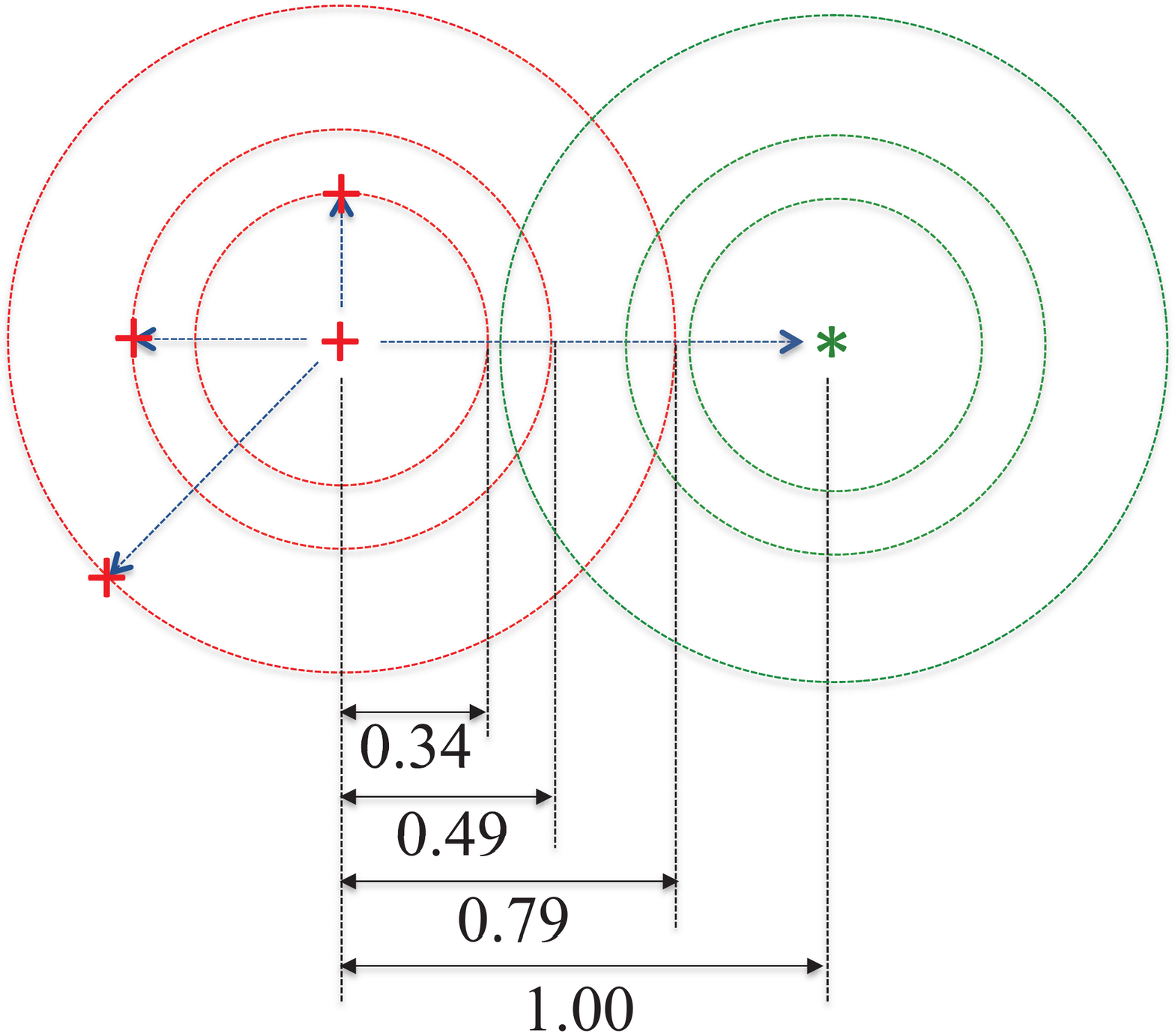} \\ [-1mm]
(a) & (b) \\ [-1mm]
\end{tabular}
\end{center}
\figvspace
\caption{(a) Center estimation error comparison. (b) Illustration of intra- and inter-class variances. Circles from small to large show the minimum, mean and maximum distances from intra-class samples to center. Distances are averaged across $1$K classes.}
\label{fig:cent_err}
\vspace{-3mm}
\end{figure}

\subsection{Feature Center Estimation}
Feature center estimation is a key step for feature transfer. To evaluate center estimation for \textit{UR} classes, $1$K regular classes are selected from MS-Celeb-1M and features are extracted using a pre-trained recognition model. We randomly choose a subset of $1$, $5$, $10$, $20$ images to mimic an \textit{UR} class. Three methods are compared: (1) ``PickOne'', randomly pick one sample as center. (2) ``AvgAll'', average features of all images. (3) ``AvgFlip'', proposed method in Eqn.~\ref{eq:center}. We compute the error as the difference between the center of the full set (ground truth) and the subset (estimated), and is normalized by the inter-class variance. 

Results in Figure~\ref{fig:cent_err} show that our ``AvgFlip" achieves a smaller error. When compared to the intra-class variance, the error is fairly small, which suggests that our center estimation is accurate to support the feature transfer.

\subsection{Effects of m-$L_2$ Regularization}
To study the effects of the proposed m-$L_2$ regularization, we show a toy example on the MNIST dataset~\cite{lecun1998mnist}. We use LeNet++ network (following ~\cite{wen2016discriminative}) to learn a $2$D feature space for better visualization. Two models are compared: one trained with softmax loss only; the other trained with softmax loss and m-$L_2$ regularization ($\alpha_{reg} = 0.001$). 
\begin{figure}[t]
\begin{center}
\begin{tabular}{@{}c@{}c@{}}
\includegraphics[trim=10 0 20 0, clip, width=0.24\textwidth]{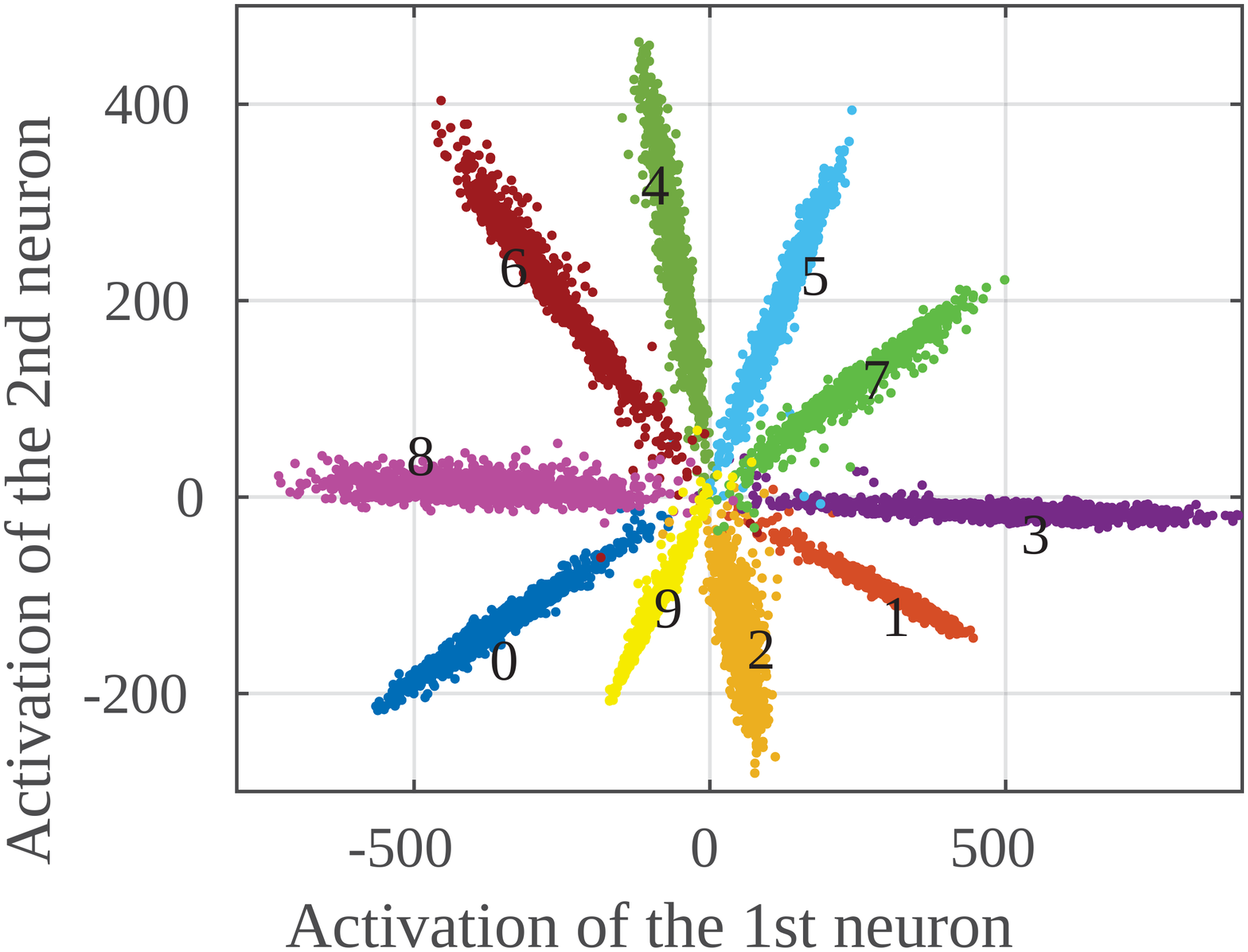} &
\includegraphics[trim=10 0 20 0, clip, width=0.24\textwidth]{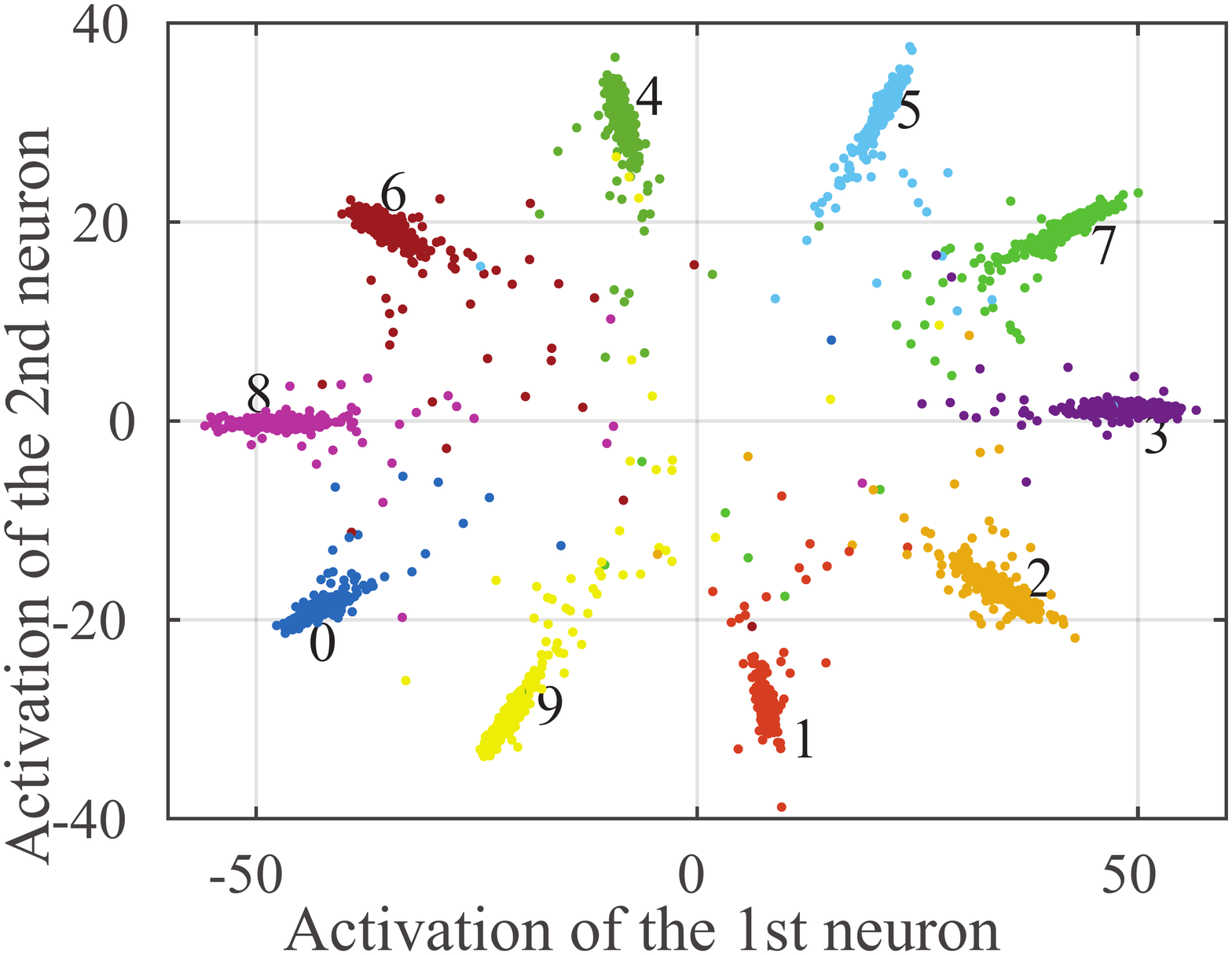} \\ [-1mm]
\small (a) & (b)  \\ [-1mm]
\end{tabular}
\end{center}
\figvspace
\caption{Toy example on MNIST to show the effectiveness of our m-$L_2$ regularization. Figure shows the feature distributions for models trained without (a) and with (b) m-$L_2$ regularization.}
\label{fig:ml2}
\vspace{-3mm}
\end{figure}

We have the following observations: (1) m-$L_2$ effectively avoids over-fitting. In Figure~\ref{fig:ml2}, the norm of the features in (a) is much larger than that in (b), as increasing the feature norm can reduce softmax loss, which may cause over-fitting. (2) m-$L_2$ enforces a more balanced feature distribution, where Figure~\ref{fig:ml2} (b) shows a more balanced angular distribution than that in (a). On the MNIST testing set, the performance with m-$L_2$ improves $sfmx$ from $99.06\%$ to $99.35\%$. Moreover, the testing accuracy with m-$L_2$ improves $sfmx$ and $sfmx+L_2$ from $98.60\%$ and $98.53\%$ to $99.37\%$ on LFW as in Table~\ref{tab:LFW}. Note that m-$L_2$ is a general regularization which is orthogonal to our main claim in this paper, that can be easily adapted to other recognition frameworks.

\begin{table*}[t]
\begin{center}
\small
\begin{tabular}{@{\hskip 3mm}l@{\hskip 2mm} |@{\hskip 5mm} l |@{\hskip 5mm} c@{\hskip 5mm}c@{\hskip 5mm}|@{\hskip 5mm}c@{\hskip 5mm}c@{\hskip 5mm}|@{\hskip 5mm}c@{\hskip 5mm}c@{\hskip 5mm}|@{\hskip 5mm}c@{\hskip 5mm}c@{\hskip 3mm}}
\hline
\multicolumn{2}{c|}{Test $\to$} & \multicolumn{2}{c|}{LFW} & \multicolumn{2}{c|}{IJB-A: Verif.} & \multicolumn{2}{c|}{IJB-A: Identif.} & \multicolumn{2}{c}{MS1M: NN} \\ \hline 
\multicolumn{1}{c|}{Train$\downarrow$} & \multicolumn{1}{c|}{Method$\downarrow$} & $\bf{g}$ &${\bf{f}}$ & FAR@$.01$ &@$.001$ & Rank-$1$ & Rank-$5$ & Reg. & \textit{UR} \\ \hline \hline  
\multirow{2}{*}{$10$K$0$K} & sfmx & $97.15$ & $97.45$ & $69.39$ & $33.04$ & $81.63$ & $90.35$ & $87.17$ & $82.47$\\
& sfmx+m-$L_2$ & $97.00$ & $97.88$ & $73.00$ & $44.78$ & $83.77$ & $91.49$ & $90.21$ & $84.68$ \\ \hline 
\multirow{3}{*}{$10$K$10$K} & sfmx & \textendash & $97.85$ & $72.96$ & $49.22$ & $82.38$ & $90.46$ & $85.87$ & $85.25$\\
& sfmx+m-$L_2$ & $97.08$ & $97.85$ & $74.07$ & $46.27$ & $83.70$ & $91.74$ & $89.48$ & $84.10$ \\ 
& FTL (Ours)$^*$ & $96.72$ & $98.33$ & $80.25$ & $54.95$ & $85.88$ & $92.83$ & $92.27$ & $88.16$ \\ \hline  
\multirow{3}{*}{$10$K$30$K} & sfmx & \textendash & $97.80$ & $74.03$ & $47.93$ & $83.04$ & $91.25$ & $86.14$ & $85.47$\\
& sfmx+m-$L_2$ & $97.13$ & $98.08$ & $76.92$ & $47.17$ & $84.81$ & $91.93$ & $90.60$ & $86.40$ \\ 
& \xiyin{FTL (Ours)$^*$} & $96.87$ & $98.42$ & $81.80$ & $61.04$ & $86.08$ & $92.62$ & $91.76$ & $88.72$ \\ \hline 
\multirow{3}{*}{$10$K$50$K} & sfmx & \textendash & $97.93$ & $72.87$ & $49.04$ & $82.40$ & $91.15$ & $85.28$ & $84.21$\\
& sfmx+m-$L_2$ & $97.32$ &$98.10$ & $78.52$ & $53.44$ & $84.95$ & $92.17$ & $90.24$ & $87.11$ \\ 
& \xiyin{FTL (Ours)$^*$} & $96.95$ &$98.48$ & $82.60$ & $62.60$ & $86.53$ & $93.08$ & $92.08$ & $89.36$ \\ \hline 
\multirow{2}{*}{$60$K$0$K} & sfmx & $97.52$ & $98.30$ & $82.75$ & $62.33$ & $87.11$ & $93.78$ & $90.43$ & $89.54$ \\ 
& sfmx+m-$L2$ & $\bf 97.90$ & $\bf 98.85$ & $\bf 86.38$ & $\bf 74.44$ & $\bf 89.34$ & $\bf 94.65$ & $\bf 93.68$ & $\bf 93.46$ \\ \hline  
\end{tabular}
\end{center}
\figvspace
\caption{Controlled experiments by varying the ratio between regular and \textit{UR} classes in training sets. FTL (Ours)$^*$: model trained on subsets.}
\label{tab:control_ratio}
\vspace{-3mm}
\end{table*}

\subsection{Ablation Study}
We study the impact of the ratio between the portion of regular classes and the portion of \textit{UR} classes on training a face recognition system. To construct the exact regular and \textit{UR} classes, we use the top $60$K regular classes, which contain the most images from MS-Celeb-1M. Further, the top $10$K classes are selected as regular classes which are shared among all training sets. We regard the $10$K and $60$K sets as the lower and upper bounds. Among the rest $50$K classes sorted by the number of images, we select the first $10$K, $30$K and $50$K and randomly pick $5$ images per class. In this way, we form the training set of $10$K$10$K, $10$K$30$K, and $10$K$50$K, of which the first $10$K are regular and the last $10$K or $30$K or $50$K are called faked \textit{UR} classes. A hold-out testing set is formed by selecting $5$ images from each of the shared $10$K regular classes and $10$K \textit{UR} classes.

The evaluation on the hold out test set from MS-Celeb-1M is to mimic low-shot learning, where we use the feature center from the training images as the gallery and nearest neighbor (NN) for face matching. The rank-$1$ accuracy for both regular and \textit{UR} classes are reported. We also evaluate the recognition performance on LFW and IJB-A. The results are shown in Table~\ref{tab:control_ratio} and we draw the following observations.
\vspace{-6mm}
\begin{tight_itemize}
\item{The rich feature space $\bf{g}$ is less discriminative than the feature space ${\bf{f}}$, which validates our intuition that $\bf{g}$ is rich in intra-class variance for feature transfer while ${\bf{f}}$ is more discriminative for face recognition.}
\item{The proposed m-$L_2$ regularization boosts the performance with a large margin over the baseline softmax loss.}
\item{The proposed FTL method consistently improves over softmax and sfmx+m-$L_{2}$ with significant margins.}
\item{Our method is more beneficial when more \textit{UR} classes are used for training as more training data usually lead to better face recognition performance.}
\end{tight_itemize}

\begin{table}[t]
\begin{center}
\small
\begin{tabular}{@{\hskip 2mm}l|@{\hskip 2mm}c|@{\hskip 2mm}c|@{\hskip 2mm}c|@{\hskip 2mm}c@{\hskip 2mm}}
\hline
Method & Ext & $\#$Models &Base & Novel \\ \hline \hline
MCSM~\cite{xu2017high} & YES & $3$ & $\textendash$ & $61.0$ \\
Cheng et al.~\cite{cheng2017know} & YES & $4$ & $99.74$ & $\bf 100$ \\ \hline
Choe et al.~\cite{choe2017face} & NO & $1$ &$\ge95.00$ & $11.17$ \\
UP~\cite{guo2017one} & NO & $1$ & $99.80$ & $77.48$ \\ 
Hybrid~\cite{wu2017low} & NO & $2$ & $99.58$ & $\bf 92.64$ \\ 
DM~\cite{smirnov2017doppelganger} & NO & $1$ & $\textendash$ & $73.86$ \\ \hline
\xiyin{FTL (Ours)} & NO & $1$ &$99.21$ & $92.60$ \\ \hline 
\end{tabular}
\end{center}
\figvspace
\caption{Comparison on one-shot learning challenge. Result on base classes are reported as rank-$1$ accuracy and on novel classes as Coverage@Precision = $0.99$. ``Ext'' means ``External Data''.}
\label{tab:lowshot}
\vspace{0mm}
\end{table}

\begin{table}
\centering
\small
\begin{tabular}{@{\hskip 2mm}l@{\hskip 3mm}|@{\hskip 3mm}c|@{\hskip 3mm}l|@{\hskip 3mm}c@{\hskip 2mm}}
\hline
Method & Acc  & Method & Acc \\ \hline \hline
L-Softmax~\cite{liu2016large} & $98.71$ & \xiyin{ArcFace~\cite{deng2019arcface}} & \xiyin{$99.53$}  \\
VGG Face~\cite{ParkhiVGG} & $98.95$ & FaceNet~\cite{schroff2015facenet}& ${\bf99.63}$\\
DeepID2~\cite{Sunnips2014} & $99.15$ & CosFace~\cite{wang2018cosface} & $\bf 99.73$\\
NormFace~\cite{wang2017normface} & $99.19$ & sfmx & $98.60$\\
CenterLoss~\cite{wen2016discriminative} & $99.28$ & sfmx + $L_2$ & $98.53$\\
SphereFace~\cite{liu2017sphereface} & $99.42$ & sfmx + m-$L_2$ (Ours) & $99.18$ \\
RangeLoss~\cite{zhang2017range} & $99.53$ & \xiyin{FTL (Ours)} & $99.55$\\\hline 
\end{tabular}
\vspace{1mm}
\caption{Performance comparisons on LFW. Methods of sfmx, sfmx+$L_2$, sfmx+m-$L_2$ are our implementations.}
\label{tab:LFW}
\vspace{0mm}
\end{table}

\begin{table}[t]
\small
\centering
\begin{tabular}{@{\hskip 0.5mm}l@{\hskip 2.5mm}|@{\hskip 2.5mm}c@{\hskip 2.5mm}c@{\hskip 2.5mm}|@{\hskip 2.5mm}c@{\hskip 2.5mm}c@{\hskip 2.5mm}c@{\hskip 0.5mm}}
\hline
Test $\to$ &\multicolumn{2}{c|}{Verification} &\multicolumn{3}{c}{Identification}\\ \hline 
Method $\downarrow$ & $0.01$ & $0.001$ & $1$ & $5$ & $10$\\ \hline \hline
PAMs~\cite{MasiCVPR16} & $82.6$ & $65.2$ & $84.0$ & $92.5$ & $94.6$ \\
DR-GAN~\cite{tran2017} & $83.1$ & $69.9$ & $90.1$ & $95.3$ & $\textendash$ \\
FF-GAN~\cite{yin2017towards} & $85.2$ & $66.3$ & $90.2$ & $95.4$ & $\textendash$\\ 
TA~\cite{crosswhite2017template} &$93.9$ & $\textendash$ & $92.8$ & $\textendash$ & $98.6$ \\
TPE~\cite{sankaranarayanan2016triplet} & $90.0$ & $81.3$ & $86.3$ & $93.2$ & $97.7$ \\
NAN~\cite{yang2016neural} & $94.1$ & $88.1$ & $95.8$ & $98.0$ & $98.6$\\ \hline
sfmx & $91.5$ & $77.4$ & $92.4$ & $96.4$ & $97.3$ \\
sfmx + m-$L_2$ (Ours) & $92.5$ & $80.2$ & $93.9$ & $97.2$ & $97.9$ \\
\xiyin{FTL (Ours)} & $93.5$ & $82.9$ & $94.8$ & $97.8$ & $98.3$ \\ 
FTL + MP (Ours) & $94.3$ & $85.1$ & $95.1$ & $97.8$ & $98.4$ \\
FTL + MP + TA (Ours) & $\bf 95.3$ & $\bf 91.2$ & $\bf 96.0$ & $\bf 98.3$ & $\bf 98.7$ \\ \hline
\end{tabular}
\vspace{1mm}
\caption{Face recognition results on IJB-A. ``MP'' and ``TA'' represent media pooling and template adaptation. Verification and identification results are reported at different FARs and ranks.}
\label{tab:large-scale}
\vspace{-1mm}
\end{table}

\begin{figure}
\begin{center}
\begin{tabular}{@{}c@{\hskip 0.1mm}c@{\hskip 0.1mm}c@{\hskip 0.1mm}c@{\hskip 0.1mm}c@{\hskip 0.1mm}c@{\hskip 0.1mm}c@{\hskip 0.1mm}c@{\hskip 0.1mm}c@{\hskip 0.1mm}c@{\hskip 0.1mm}c@{\hskip 0.1mm}c@{\hskip 0.1mm}c@{}}
 (a) &
 \includegraphics[width=0.07\textwidth]{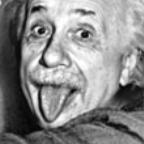} &
 \includegraphics[width=0.07\textwidth]{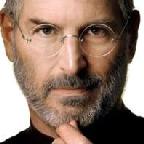} &
 \includegraphics[width=0.07\textwidth]{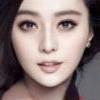} &
 \includegraphics[width=0.07\textwidth]{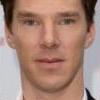} &
 \includegraphics[width=0.07\textwidth]{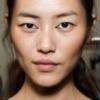} &
 \includegraphics[width=0.07\textwidth]{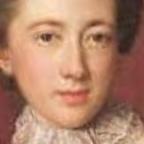} \\ [-1mm]
 (b) &
 \includegraphics[width=0.07\textwidth]{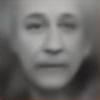} &
 \includegraphics[width=0.07\textwidth]{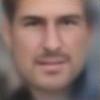} &
 \includegraphics[width=0.07\textwidth]{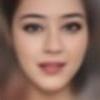} &
 \includegraphics[width=0.07\textwidth]{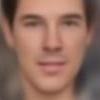} &
 \includegraphics[width=0.07\textwidth]{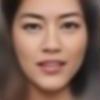} &
 \includegraphics[width=0.07\textwidth]{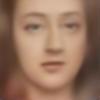} \\ [-2mm]
 \end{tabular}
 \end{center}
 \figvspace
 \caption{Center visualization. (a) one sample image from the selected class; (b) the decoded image from the feature center.}
 \label{fig:center_visual}
 \vspace{-2mm}
\end{figure} 

\secvspace
\subsection{One-Shot Face Recognition}
As our method has tangential relation to low-shot learning, we evaluate on the MS-celeb-1M one-shot challenge~\cite{guo2017one}. The training data consists of a base set with $20$K classes each with $50$$\sim$$100$ images and a novel set of $1$K classes each with only $1$ image. The test set consists of $1$ image per base (regular) class and $5$ images per novel (\textit{UR}) class. The goal is to evaluate the performance on the novel classes while monitoring the performance on base classes. 

We use the output from softmax layer as the confidence score and achieve $92.60\%$ coverage at precision of $0.99$ with single-model single-crop testing, as in Table~\ref{tab:lowshot}. Note that both methods~\cite{cheng2017know, wu2017low} use model ensemble and multi-crop testing. Compared to methods~\cite{guo2017one,choe2017face} with similar setting, we achieve competitive performance on the base classes and much better accuracy on the novel classes by $15\%$.

\begin{figure*}
\centering
\includegraphics[trim=5 200 50 195, clip,width=0.99\textwidth]{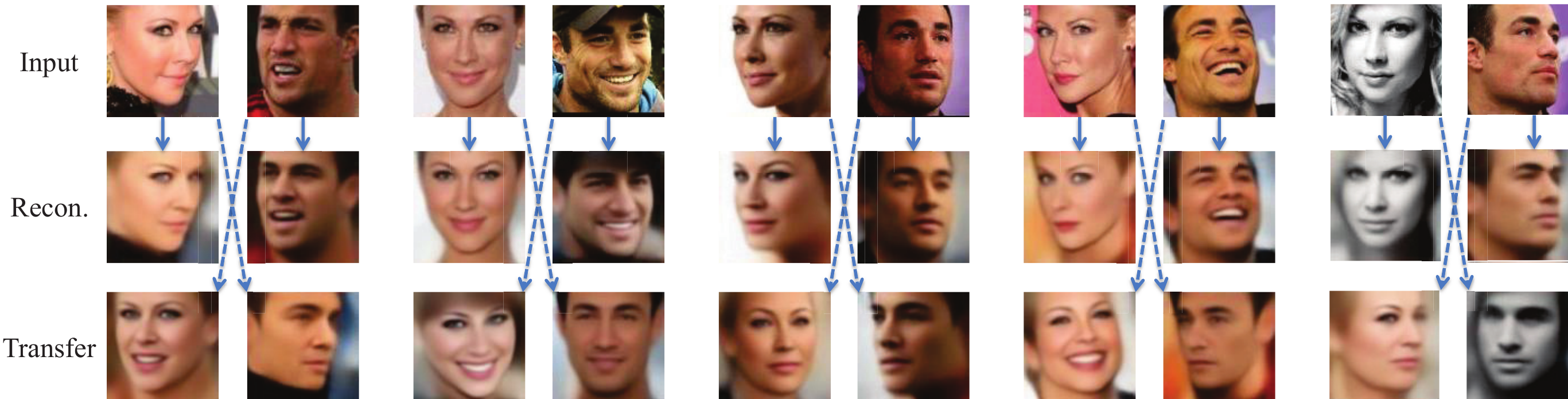}
\vspace{0mm}
\caption{Feature transfer visualization between two classes for every two columns. The first row are the input, in which odd column denotes class $1$: ${\bf{x}}_1$ and the even column denotes class $2$: ${\bf{x}}_2$. The second row are the reconstructed images ${\bf{x}}'_1$ and ${\bf{x}}'_2$. In the third row, odd column image is the decoded image of the transferred feature from class $1$ to class $2$ and even column image is the decoded image of the transferred feature from class $2$ to class $1$. It is clear that the transferred features share the same identity as the target class while obtain the source image's non-identity variance including pose, expression, illumination, and \etc.}
\label{fig:transfer}
\vspace{-1mm}
\end{figure*}

\begin{figure*}[t]
\begin{center}
\begin{tabular}{@{}c@{\hskip 0.8mm}c@{\hskip 0.2mm}c@{\hskip 0.2mm}c@{\hskip 0.2mm}c@{\hskip 0.2mm}c@{\hskip 0.2mm}c@{\hskip 0.2mm}c@{\hskip 0.2mm}c@{\hskip 0.2mm}c@{\hskip 0.2mm}c@{\hskip 0.8mm}c@{}}
\includegraphics[trim=0 201 1100 0, clip, width=0.078\textwidth]{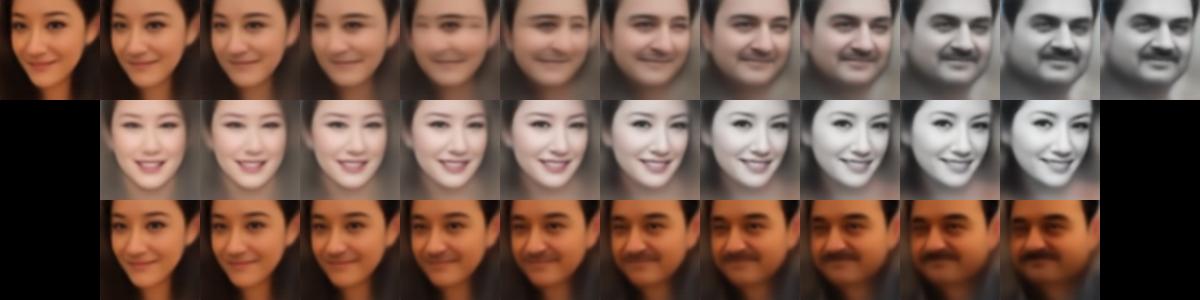} &
\includegraphics[trim=100 201 1000 0, clip, width=0.078\textwidth]{fig/interpolation_temp_792.jpg} &
\includegraphics[trim=200 201 900 0, clip, width=0.078\textwidth]{fig/interpolation_temp_792.jpg} &
\includegraphics[trim=300 201 800 0, clip, width=0.078\textwidth]{fig/interpolation_temp_792.jpg} &
\includegraphics[trim=400 201 700 0, clip, width=0.078\textwidth]{fig/interpolation_temp_792.jpg} &
\includegraphics[trim=500 201 600 0, clip, width=0.078\textwidth]{fig/interpolation_temp_792.jpg} &
\includegraphics[trim=600 201 500 0, clip, width=0.078\textwidth]{fig/interpolation_temp_792.jpg} &
\includegraphics[trim=700 201 400 0, clip, width=0.078\textwidth]{fig/interpolation_temp_792.jpg} &
\includegraphics[trim=800 201 300 0, clip, width=0.078\textwidth]{fig/interpolation_temp_792.jpg} &
\includegraphics[trim=900 201 200 0, clip, width=0.078\textwidth]{fig/interpolation_temp_792.jpg} &
\includegraphics[trim=1000 201 100 0, clip, width=0.078\textwidth]{fig/interpolation_temp_792.jpg} &
\includegraphics[trim=1100 201 0 0, clip, width=0.078\textwidth]{fig/interpolation_temp_792.jpg} \\ [-1mm]
&
\includegraphics[trim=100 100 1000 101, clip, width=0.078\textwidth]{fig/interpolation_temp_792.jpg} &
\includegraphics[trim=200 100 900 101, clip, width=0.078\textwidth]{fig/interpolation_temp_792.jpg} &
\includegraphics[trim=300 100 800 101, clip, width=0.078\textwidth]{fig/interpolation_temp_792.jpg} &
\includegraphics[trim=400 100 700 101, clip, width=0.078\textwidth]{fig/interpolation_temp_792.jpg} &
\includegraphics[trim=500 100 600 101, clip, width=0.078\textwidth]{fig/interpolation_temp_792.jpg} &
\includegraphics[trim=600 100 500 101, clip, width=0.078\textwidth]{fig/interpolation_temp_792.jpg} &
\includegraphics[trim=700 100 400 101, clip, width=0.078\textwidth]{fig/interpolation_temp_792.jpg} &
\includegraphics[trim=800 100 300 101, clip, width=0.078\textwidth]{fig/interpolation_temp_792.jpg} &
\includegraphics[trim=900 100 200 101, clip, width=0.078\textwidth]{fig/interpolation_temp_792.jpg} &
\includegraphics[trim=1000 100 100 101, clip, width=0.078\textwidth]{fig/interpolation_temp_792.jpg} \\[-1mm]
&
\includegraphics[trim=100 0 1000 201, clip, width=0.078\textwidth]{fig/interpolation_temp_792.jpg} &
\includegraphics[trim=200 0 900 201, clip, width=0.078\textwidth]{fig/interpolation_temp_792.jpg} &
\includegraphics[trim=300 0 800 201, clip, width=0.078\textwidth]{fig/interpolation_temp_792.jpg} &
\includegraphics[trim=400 0 700 201, clip, width=0.078\textwidth]{fig/interpolation_temp_792.jpg} &
\includegraphics[trim=500 0 600 201, clip, width=0.078\textwidth]{fig/interpolation_temp_792.jpg} &
\includegraphics[trim=600 0 500 201, clip, width=0.078\textwidth]{fig/interpolation_temp_792.jpg} &
\includegraphics[trim=700 0 400 201, clip, width=0.078\textwidth]{fig/interpolation_temp_792.jpg} &
\includegraphics[trim=800 0 300 201, clip, width=0.078\textwidth]{fig/interpolation_temp_792.jpg} &
\includegraphics[trim=900 0 200 201, clip, width=0.078\textwidth]{fig/interpolation_temp_792.jpg} &
\includegraphics[trim=1000 0 100 201, clip, width=0.078\textwidth]{fig/interpolation_temp_792.jpg} \\ [-1mm]
$\alpha\to$ & $0.1$ & $0.2$ & $0.3$ & $0.4$ & $0.5$ & $0.6$ & $0.7$ & $0.8$ & $0.9$ & $1.0$ \\ 
\end{tabular}
\end{center}
\figvspace
\caption{Transition from top-left image to top-right image via feature interpolation. First row shows traditional feature interpolation; second row shows our transition of non-identity variance; third row shows our transition of identity variance.}
\label{fig:interpolation}
\vspace{-2mm}
\end{figure*}

\subsection{Large-Scale Face Recognition}
In this section, we train our model on the full MS-celeb-1M dataset and evaluate on LFW and IJB-A.
On LFW (Table~\ref{tab:LFW}), our performance is strongly competitive, achieving $99.55\%$ whereas the state-of-the-arts show $99.63\%$ from FaceNet~\cite{schroff2015facenet} and $99.73\%$ from CosFace~\cite{wang2018cosface}.
On IJB-A (Table~\ref{tab:large-scale}), the softmax loss with our proposed m-$L_2$ regularization already provides good results denoted as sfmx+m-$L_2$. Our FTL improves the performance significantly, with margins varying from $0.6\%$ to $2.8\%$. 
We further combine media pooling (MP) and template adaptation (TA)~\cite{crosswhite2017template} metric learning with our proposed method (FTL + MP + TA), and achieve consistently better results than state-of-the-art methods~\cite{yang2016neural}. 

\subsection{Qualitative Results}
We apply decoder $Dec$ in our framework for feature visualization. 
While skip link between encoder and decoder improves the visual quality~\cite{yin2017towards}, we do not apply it to encourage the rich features $\bf{g}$ to encode intra-class variance.

\Paragraph{Center visualization}
We compute a feature center for a given class, on which the $Dec$ is applied to generate a center face. As shown in Figure~\ref{fig:center_visual}, we confirm the observation that the center is mostly an identity-preserved frontal neutral face. It also applies to portrait and cartoon figures. 

\Paragraph{Feature transfer}
The transferred features are visualized by $Dec$. 
Let ${\bf{x}}_{1,2}$, ${\bf{x}}'_{1,2}$, ${\bf{g}}_{1,2}$, ${\bf{c}}_{1,2}$ denote the input images, reconstructed images, encoded rich features and feature centers of two classes, respectively. 
We transfer feature from class $1$ to class $2$ by: ${\bf{g}}_{12} = {\bf{c}}_2 + {\bf{Q}}{\bf{Q}}^{T}({\bf{g}}_1 - {\bf{c}}_1)$, and visualize the decoded images. 
We also transfer from class $2$ to class $1$ and visualize the decoded images.
As shown in Figure~\ref{fig:transfer}, the transferred images preserve the target class's identity while retaining intra-class variance of the source image in terms of pose, expression and lighting, which shows that our feature transfer is effective in enlarging the intra-class variance.

\Paragraph{Feature interpolation}
The interpolation between two representations shows the appearance transition from one to the other~\cite{radford2015unsupervised, tran2017}.
Let ${\bf{g}}_{1,2}$, ${\bf{c}}_{1,2}$ denote the encoded features and the centers of two classes.
Previous work generates a new representation as ${\bf{g}} = {\bf{g}}_1 + \alpha({\bf{g}}_2 - {\bf{g}}_1)$ where identity and non-identity changes are mixed together. 
In our work, we can generate transitions of non-identity change as ${\bf{g}} = {\bf{c}}_1 +\alpha {\bf{Q}}{\bf{Q}}^{T}({\bf{g}}_2 - {\bf{c}}_2)$ and identity change as ${\bf{g}} = {\bf{g}}_1 + \alpha({\bf{c}}_2 - {\bf{c}}_1)$.
Figure~\ref{fig:interpolation} shows an interpolation example of a female with left pose and a male with right pose, where the illumination changes significantly. 
Compared to traditional interpolation that generates undesirable artifacts, our method shows smooth transitions, which verifies that the proposed model is effective at disentangling identity and non-identity features. 

\secvspace
\section{Conclusions}
\vspace{-2mm}
In this paper, we propose a novel feature transfer approach for deep face recognition training which explores the imbalance issue with {\textit{UR}} classes.
We observe that generic face recognition approaches encounter classifier bias due to imbalanced distribution of training data across classes.
By applying the proposed feature transfer approach, we enrich the feature space of the \textit{UR} classes, while retaining identity.
Utilizing the generated data, our alternating feature learning method rectifies the classifier and learns more compact feature representations. Our proposed m-$L_2$ regularization demonstrates consistent advantages which can potentially boost performance across different recognition tasks. The disentangled nature of the augmented feature space is visualized through smooth interpolations.
Experiments consistently show that our method can learn better representations to improve the performance on regular, \textit{UR}, and unseen classes.
While this paper focuses on face recognition, our future work will also derive advantages from the proposed feature transfer for other recognition applications, such as \textit{UR} natural species~\cite{fgvc4}.

{\small
\bibliographystyle{ieee}
\bibliography{egbib}
}

\end{document}